\documentclass[twoside,11pt]{article}

\usepackage{automl2020}

\usepackage{overpic}
\usepackage{wrapfig}

\usepackage{amsmath}
\usepackage{amssymb}
\usepackage{mathrsfs}

\usepackage{enumitem}

\usepackage{xcolor}
\usepackage{arydshln}

\usepackage{CJKutf8}
\usepackage{ucs}
\usepackage[encapsulated]{CJK} 

\usepackage{algorithmicx}
\usepackage{algorithm2e}
\usepackage{caption}
\usepackage{xcolor}
\definecolor{Q-col}{rgb}{1,0.835,0.835}
\definecolor{S-col}{rgb}{0.835,0.835,1}
\definecolor{Xi-col}{rgb}{0.867,1,0.835}
\definecolor{Theta-col}{rgb}{0.835,1,1}
\definecolor{Phi-col}{rgb}{1,1,0.835}

\usepackage{subcaption}

\hypersetup{
  colorlinks,
  citecolor=green,
  linkcolor=red}

\jmlrheading{Nicholas Kuo, Mehrtash Harandi, Nicolas Fourrier, Christian Walder, Gabriela Ferraro, and Hanna Suominen}

\ShortHeadings{MTL2L: A Context Aware Neural Optimiser}{Kuo et al.}
\firstpageno{1}

\begin{document}

\title{MTL2L: A Context Aware Neural Optimiser}

\author{\name Nicholas I-Hsien Kuo$^1$ \email u6424547@anu.edu.au \\
       \name Mehrtash Harandi$^2$ \email mehrtash.harandi@monash.edu \\
       \name Nicolas Fourrier$^3$ \email nfourrier@gmail.com \\
       \name Christian Walder$^{1, 4}$ \email christian.walder@data61.csiro.au \\
       \name Gabriela Ferraro$^{1, 4}$ \email gabriela.ferraro@data61.csiro.au \\
       \name Hanna Suominen$^{1, 4, 5}$ \email hanna.suominen@anu.edu.au \\
       \addr 
       $^1:$ RSCS, The Australian National University, Canberra, ACT, Australia\\
       $^2:$ ECSE, Monash University, Melbourne, Victoria, Australia\\
       $^3:$ Research Center of L\'{e}onard de Vinci P\^{o}le Universitaire, Paris La D\'{e}fense, France\\
       $^4:$ Data61, CSIRO, Canberra, ACT, Australia\\
       $^5:$ Department of Future Technologies, University of Turku, Turku, Finland}

\maketitle

\begin{abstract}
Learning to learn (L2L) trains a meta-learner to assist the learning of a task-specific base learner. Previously, it was shown that a meta-learner could learn the direct rules to update learner parameters; and that the learnt neural optimiser updated learners more rapidly than handcrafted gradient-descent methods. However, we demonstrate that previous neural optimisers were limited to update learners on \textit{one} designated dataset. In order to address input-domain heterogeneity, we introduce \textbf{Multi-Task Learning to Learn (MTL2L)}, a context aware neural optimiser which self-modifies its optimisation rules based on input data. We show that MTL2L is capable of updating learners to classify on data of an unseen input-domain at the meta-testing phase.\newline
\underline{We release our codes at} \url{https://github.com/Nic5472K/AutoML2020_ICML_MTL2L}.\\
\end{abstract}

\section{Introduction}\label{sec:1}

Deep learning has firmly established itself in image recognition \citep{he2016deep}, speech recognition \citep{oord2016wavenet}, and text processing \citep{vaswani2017attention} through training models with high-capacity on a large amount of labelled data. However, it remains difficult to learn from a low resource environment. This challenging task is carefully studied in meta-learning, the branch of machine learning that addresses rapid knowledge acquisition from only a few examples. In this paper, we address extensions to the neural optimiser approach for the meta-learning task of \textit{Learning to learn} (L2L)~\citep{andrychowicz2016learning}.

L2L formulates the concept of learning quickly as the task to rapidly converge a network with the least amount of iterations. This task involves training a meta-learner to assist the learning of a base learner. The learner performs a specific task, such as image classification; whereas the meta-learner captures how task structures vary within the domain \citep{rendell1987layered}. This paper extends on meta-learners which directly update learner parameters.

\citet{andrychowicz2016learning} cast optimisation as the learning target for a recurrent neural network (RNN). The RNN was \textit{meta-trained} as a neural optimiser meta-learner; then \textit{meta-tested} to update a base learner. They showed that the \textit{RNN neural optimiser} converged 

\newpage
\begin{figure}[t!]
\begin{center}
  \centering
  \label{Fig:Intro-F1}
  \begin{overpic}[width=11.34cm,height=3.9299904cm]{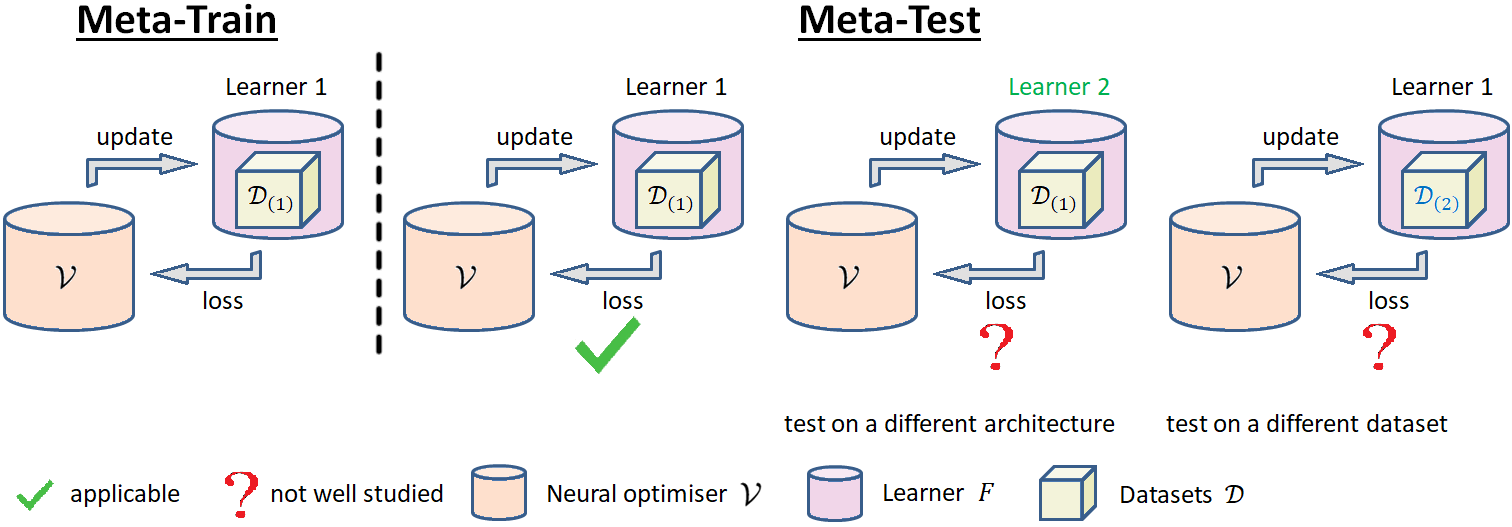}
  
  \end{overpic}
\end{center}
\vspace{-5mm}
\caption{Neural optimiser applicability}
\end{figure}

\vspace*{-8mm}
\begin{wrapfigure}{r}{.375\textwidth}
\label{Fig:Intro-F2}
    \begin{minipage}{\linewidth}
    \centering
    \vspace*{-5mm}
    \includegraphics[width=0.9\linewidth]{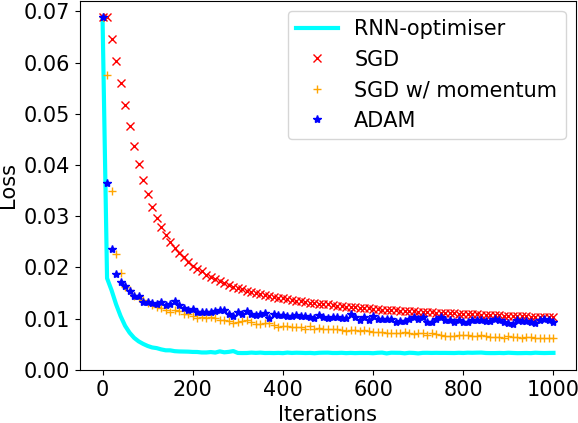}
    \vspace*{-2mm}
    \subcaption*{\hspace{5mm}\footnotesize{(a) Meta-trained on MNIST and \\
      \vspace{-2mm}\hspace{10.25mm}Meta-tested on MNIST}}
    \label{fig:2a}\par\vfill
    \vspace*{2mm}
    \includegraphics[width=0.9\linewidth]{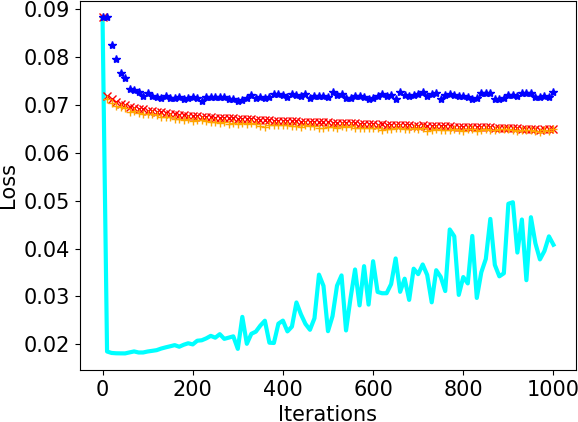}
    \vspace*{-2mm}
    \subcaption*{\hspace{5mm}\footnotesize{(b) Meta-trained on MNIST and \\
      \vspace{-2mm}\hspace{10.375mm}Meta-tested on Modified Cifar10}}
    \label{fig:2b}
\end{minipage}
\vspace{-1mm}
\caption{Domain variance}
\vspace{-3mm}
\end{wrapfigure}

\hspace{-8mm} learners more rapidly than hand-designed optimisations such as SGD~\citep{robbins1951stochastic} and ADAM~\citep{kingma2014adam}. Learning to optimise has since become a common approach in meta-learning, and has been extended by the LSTM-metalearner~\citep{ravi2016optimization} and Meta-SGD~\citep{li2017meta}.

However, neural optimisers have the two limitations\footnote{See more descriptions in \textbf{Appendix }\hyperref[AppA]{\textbf{A}}.} illustrated in \textbf{Figure }\hyperref[Fig:Intro-F1]{\textbf{1}}. First, neural optimisers are specific to architectures. If they were meta-trained to update a multiple layer perceptron (MLP), it cannot be guaranteed to work on ResNet~\citep{he2016deep}. Second, neural optimisers are specific to input-domain (dataset). When naively applied, they cannot update the same learner on two distinct datasets as the underlying task structures may vary significantly in the two input-domains. Neural optimisers are thus not as generally applicable as SGD.

We prepare \textbf{Figure }\hyperref[Fig:Intro-F2]{\textbf{2}} to demonstrate and to convince our readers of the consequences of naively applying neural optimisers on an unseen dataset. The solid cyan lines represent the RNN-optimiser of \citet{andrychowicz2016learning}. \textbf{Figure }\hyperref[fig:2a]{\textbf{2(a)}} illustrates meta-learning under the common assumption: \textit{to meta-test on the single identical input-domain used in meta-training}; and \textbf{Figure }\hyperref[fig:2b]{\textbf{2(b)}} depicts deteriorated performance in meta-testing on an unseen dataset. In this paper, we introduce \textbf{Multi-Task Learning to Learn (MTL2L)} -- a novel context aware neural optimiser which supports \textit{multi-task learning}~\citep{caruana1997multitask}. MTL2L can capture shared task structures across multiple input-domains to update learners on a dataset unseen during meta-training. The experiments will be revisited in \textbf{Section }\hyperref[SecExp]{\textbf{4}}.

\vspace*{-2mm}
\section{Preliminaries}
The classic gradient descent updates parameters \colorbox{Theta-col}{$\theta_t$} of learner $F$ at time $t$ via
\begin{align}
\theta_{t+1} &=\theta_t - \alpha \nabla_\theta \mathcal{L}_{t+1}(\theta_t) \label{sec2eq1}
\end{align}

\newpage
\begin{figure}[t!]
\label{sec2fig3}
\begin{center}
  \centering
  \begin{overpic}[width=9cm,height=5.346cm]{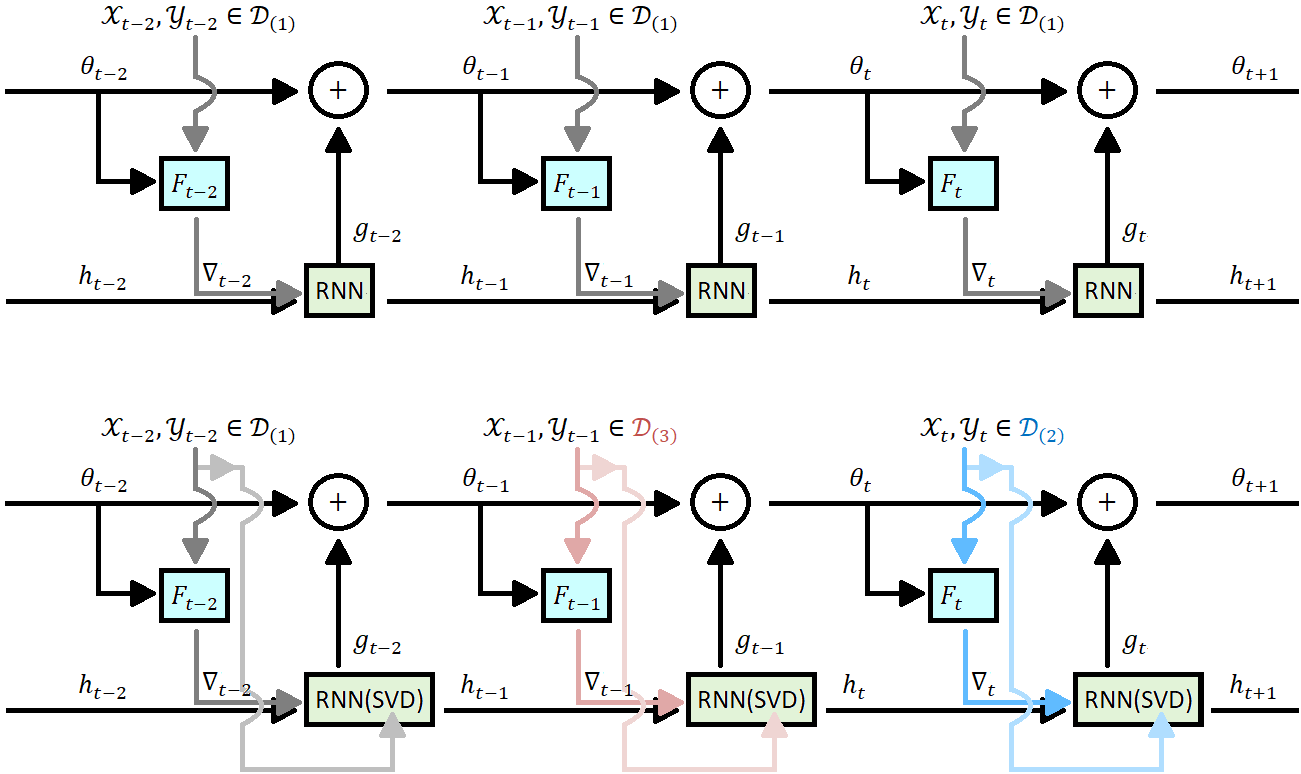}
  
  \put(32, 30.5) {\small (a) The LSTM-optimiser}
  \put(40, -4) {\small (b) MTL2L}
  
  \end{overpic}
\end{center}
\vspace*{-3mm}
\caption{A comparison between the LSTM-optimiser and MTL2L}
\end{figure}

\vspace*{-8mm}
\hspace{-6mm}with learning rate $\alpha$ and gradients of the learner loss with respect to the parameters $\nabla_\theta \mathcal{L}_{t+1}(\theta_t)$. \citet{andrychowicz2016learning} proposed to replace Equation \eqref{sec2eq1} with an RNN $\mathcal{V}$ with parameters \colorbox{Phi-col}{$\phi$} to learn the direct rules $g_t$ to update the learner parameters
\begin{align}
\theta_{t+1} &=\theta_t + g_{t+1} \hspace{2mm} \text{and} \label{sec2eq2}\\
\left[g_{t+1}, h_{t+1}\right] &= \mathcal{V}\big(\nabla_\theta\mathcal{L}_{t+1}(\theta_t), h_t \lvert \phi \big), \label{sec2eq3}
\end{align}
\noindent
\begin{minipage}{.475\linewidth}
\label{sec2alg1}
    \begin{algorithm}[H]\footnotesize
    $\tau = 0$\\
    $\phi \leftarrow \text{random initialisation}$ \\
    \For{$q = 1,\colorbox{Q-col}{$\mathcal{Q}$}$}{
        $\theta_0 \leftarrow \text{random initialisation}$
        
        \For{$t = 1,\colorbox{S-col}{$\mathcal{S}$}$}{
        $\mathcal{X}_t, \mathcal{Y}_t \leftarrow \text{sample from } \mathcal{D}_{\text{Train}}$\\
        $\mathcal{L}_{t+1}(\theta_t) \leftarrow \mathcal{L}\big(F(\mathcal{X}_t\lvert\theta_t), \mathcal{Y}_t\big)$\\
        $\tau \leftarrow \tau + 1$\\
        $\mathcal{L}_{\tau+1}(\theta_\tau) \leftarrow \mathcal{L}_{t+1}(\theta_t)$\\
        
        \If{$\tau = $\colorbox{Xi-col}{$\Xi$}}{
            $\mathscr{L}(\phi)=\mathbb{E}\left[\sum_{\tau=1}^{\Xi}\mathcal{L}_{\tau+1}(\theta_\tau)\right]$\\
            \fbox{Update meta-learner}\\
            \colorbox{Phi-col}{$\phi\leftarrow \phi - \omega\nabla_\phi\mathscr{L}(\phi)$}\\
            $\tau = 0$
            }
        
        $g_{t+1} \leftarrow
        \mathcal{V}\left(\nabla_\theta\mathcal{L}_{t+1}(\theta_t)\lvert\phi\right)$\\
        \fbox{Update learner}\\
        \colorbox{Theta-col}{$\theta_{t+1} \leftarrow \theta_t + g_t $}
        }
    }
    \caption{Meta-training}
    \end{algorithm}
\end{minipage}
\hspace{-7.5mm}
\begin{minipage}{0.5625\textwidth}
where $h_t$ is the hidden states of $\mathcal{V}$. This concept is illustrated in \textbf{Figure }\hyperref[sec2fig3]{\textbf{3(a)}}.

We now describe the training of the neural optimiser with \textbf{Algorithm }\hyperref[sec2alg1]{\textbf{1}}.
Neural optimiser $\mathcal{V}$ is trained via updating \colorbox{Q-col}{$\mathcal{Q}$} learners for \colorbox{S-col}{$\mathcal{S}$} steps on the training data $\mathcal{D}_{\text{Train}}$. 
They are updated along with the learner but on a different time scale -- only once every \colorbox{Xi-col}{``unroll''}. The objective loss $\mathscr{L}$ of $\mathcal{V}$ is dependent on the objective loss $\mathcal{L}$ of $F$ over \colorbox{Xi-col}{$\Xi$} time steps, and is defined as
\begin{align}
\mathscr{L}(\phi) &= \mathbb{E}\left[\sum_{\tau=1}^{\Xi}\mathcal{L}_{\tau+1}(\theta_\tau)\right]. \label{sec2eq4}
\end{align}
The neural optimiser is updated via classic gradient descent with learning rate $\omega$; and $\mathscr{L}$ shows that $\mathcal{V}$ is set to learn the trajectory for optimising learner $F$ with $\Xi$ units of memory bandwidth for its RNN. 
\end{minipage}

\section{Context awareness through HyperNetwork}\label{Sec3}
\citet{andrychowicz2016learning} employed the \textit{long short-term memories} (LSTMs)~\citep{hochreiter1997long, gers1999learning} presented in Equations \eqref{LSTM_eqStart} to \eqref{LSTM_eqEnd} in \textbf{Table }\hyperref[sec3tab1]{\textbf{1}} as RNNs for their neural optimiser. Symbols $\sigma$, $\tanh$, and $\odot$ are the sigmoid function, hyperbolic tangent function, and element-wise dot product, respectively; with synapses $\mathbf{W}$ and $\mathbf{U}$, and bias $\mathbf{b}$. The forget gate $\mathbf{f}_t$ and the input gate $\mathbf{i}_t$ modulate cell $\mathbf{c}_t$ with new 

\newpage
\begin{table*}[ht]
  \centering
  \label{sec3tab1}
\begin{tabular}{|c|c|}
\hline
\vspace{-1mm}
\footnotesize{\textbf{Traditional LSTM}} & \footnotesize{\textbf{MTL2L (Ours)}}\\
\hline
{
\begin{minipage}{7cm}
\begin{equation}
\footnotesize{\begin{bmatrix}
\mathbf{f}_t\\[\jot]
\mathbf{i}_t\\[\jot]
\mathbf{o}_t
\end{bmatrix}=\sigma\begin{pmatrix}
 \mathbf{W}_{\mathcal{F}} \mathbf{x}_t + \mathbf{U}_{\mathcal{F}} \mathbf{h}_{t-1} + \mathbf{b}_{\mathcal{F}}\\[\jot]
 \mathbf{W}_{\mathcal{I}} \mathbf{x}_t + \mathbf{U}_{\mathcal{I}} \mathbf{h}_{t-1} + \mathbf{b}_{\mathcal{I}}\\[\jot]
 \mathbf{W}_{\mathcal{O}} \mathbf{x}_t + \mathbf{U}_{\mathcal{O}} \mathbf{h}_{t-1} + \mathbf{b}_{\mathcal{O}} \label{LSTM_eqStart}
\end{pmatrix},}
\end{equation}
\vspace*{-1mm}
\begin{equation}
    \footnotesize{\mathbf{a}_t = \tanh(\mathbf{W}_{\mathcal{A}} \mathbf{x}_t + \mathbf{U}_{\mathcal{A}}\mathbf{h}_{t-1} + \mathbf{b}_{\mathcal{A}}) ,} 
\end{equation}
\vspace*{-5.5mm}
\begin{equation}    
    \hspace{-8.75mm}\footnotesize{\mathbf{c}_t = \mathbf{f}_t\odot\mathbf{c}_{t-1}+\mathbf{i}_t\odot\mathbf{a}_t , \hspace{2mm} \text{and}}  
\end{equation}
\vspace*{-5.5mm}
\begin{equation}    
    \hspace{-23.75mm}\footnotesize{\mathbf{h}_t = \mathbf{o}_t\odot\tanh(\mathbf{c}_t).} \label{LSTM_eqEnd}
\end{equation}
\end{minipage}} & 
{
\begin{minipage}{7cm}
\begin{equation}
\footnotesize{\begin{bmatrix}
\mathbf{f}_t\\[\jot]
\mathbf{i}_t\\[\jot]
\mathbf{o}_t
\end{bmatrix}=\sigma\begin{pmatrix}
 \textcolor{brown}{\mathcal{W}^{(\mathcal{F})}_t} \mathbf{x}_t + \textcolor{brown}{\mathcal{U}^{(\mathcal{F})}_t} \mathbf{h}_{t-1} + \mathbf{b}_{\mathcal{F}}\\[\jot]
 \textcolor{brown}{\mathcal{W}^{(\mathcal{I})}_t} \mathbf{x}_t + \textcolor{brown}{\mathcal{U}^{(\mathcal{I})}_t} \mathbf{h}_{t-1} + \mathbf{b}_{\mathcal{I}}\\[\jot]
 \textcolor{brown}{\mathcal{W}^{(\mathcal{O})}_t} \mathbf{x}_t + \textcolor{brown}{\mathcal{U}^{(\mathcal{O})}_t} \mathbf{h}_{t-1} + \mathbf{b}_{\mathcal{O}} \label{MTL2L_eqStart}
\end{pmatrix},}
\end{equation}
\vspace*{-5mm}
\begin{equation}
    \footnotesize{\mathbf{a}_t = \tanh(
    \textcolor{brown}{\mathcal{W}^{(\mathcal{A})}_t} \mathbf{x}_t + \textcolor{brown}{\mathcal{U}^{(\mathcal{A})}_t}\mathbf{h}_{t-1} + \mathbf{b}_{\mathcal{A}}) ,} 
\end{equation}
\vspace*{-7mm}
\begin{equation}    
    \hspace{-17.75mm}\footnotesize{\mathbf{c}_t = \mathbf{f}_t\odot\mathbf{c}_{t-1}+\mathbf{i}_t\odot\mathbf{a}_t , \hspace{2mm} \text{and}}  
\end{equation}
\vspace*{-5.5mm}
\begin{equation}    
    \hspace{-15.25mm}\footnotesize{\mathbf{h}_t = \mathbf{o}_t\odot\tanh(\mathbf{c}_t); \hspace{8.5mm} \text{where}}  
\end{equation}
\end{minipage}}    \\
\hdashline
\vspace{1mm}
{
\begin{minipage}{7cm}
{\footnotesize 
\textcolor{white}{.}\\
Note:\\
The modified components are coloured in \textcolor{brown}{brown}.\\
The components in \textcolor{violet}{purple} are set as \underline{constants} prior to training; whereas those in \textcolor{orange}{orange} are \underline{learnt} from training and dependent on input $x_t$.
\par}

\end{minipage}}
 & 
{
\begin{minipage}{7cm}
\begin{equation}
    \hspace{-32.75mm}\footnotesize{\mathcal{W}_t = \textcolor{violet!100}{Q} \hspace{1mm} \textcolor{orange!100}{\Gamma_t} \hspace{1mm} \textcolor{violet!100}{P}^{\text{T}} ,} \label{MTL2L_W}
\end{equation}
\vspace*{-5.5mm}
\begin{equation}
    \hspace{-6mm}\footnotesize{\mathcal{U}_t = \textcolor{violet!100}{S} \hspace{1mm} \textcolor{orange!100}{\Omega_t} \hspace{1mm} \textcolor{violet!100}{J}^{\text{T}} , \hspace{12.25mm} \text{such that}} \label{MTL2L_U}
\end{equation}
\vspace*{-5.5mm}
\begin{equation}
    \hspace{-14.25mm}\footnotesize{\Gamma_t = \text{diag}\big(\mathscr{N}^{(1)}(\mathbf{x}_t)\big) , \hspace{2mm} \text{and}} \label{HN_1}
\end{equation}
\vspace*{-5.5mm}
\begin{equation}
    \hspace{-22.5mm}\footnotesize{\Omega_t = \text{diag}\big(\mathscr{N}^{(2)}(\mathbf{x}_t)\big) .} \label{MTL2L_eqEnd}
\end{equation}

\end{minipage}}    \\
\hline
\end{tabular}
\caption{A comparison between traditional LSTM against MTL2L}
\end{table*}

\hspace{-6mm}memory $\mathbf{a}_t$. The LSTM output is hidden state $\mathbf{h}_t$, $\mathbf{c}_t$ transformed with controlled exposure from the output gate $\mathbf{o}_t$. Our novel \textbf{Multi-task Learning to Learn (MTL2L)} neural optimiser is introduced in Equations \eqref{MTL2L_eqStart} to \eqref{MTL2L_eqEnd}, and replace LSTMs as $\mathcal{V}$ in Equation \eqref{sec2eq3}.

We took a \textit{HyperNetwork}~\citep{ha2016hypernetworks} approach towards dataset-agnostic meta-learning. HyperNetworks are small auxiliary networks that re-implement weights for a larger network based on input data. Equations \eqref{MTL2L_W} and \eqref{MTL2L_U} describe MTL2L synapses as variables via \textit{singular value decomposition} (SVD). SVD splits each synapse as fixed purple terms and orange components that require learning. The former are constant orthonormal matrices, and the latter are diagonal matrices with eigenvalues as non-zero entries. 

The eigenvalues of the orange matrices are inferred with hypernetworks $\mathscr{N}=\mathscr{N}(\mathbf{x}_t)$. Since $\mathscr{N}$ yield different combinations of eigenvalues for different datasets, MTL2L becomes context aware and self-modifies its optimisation rules based on input data. It is natural for MTL2L to support multi-task learning. During meta-training, we expose MTL2L to different datasets as shown in \textbf{Figure }\hyperref[sec2fig3]{\textbf{3(b)}} -- the aim is to represent each task structure as different combinations of eigenvalues. During meta-testing, the eigenvalues enable MTL2L to compare the task structure of the input data to those learnt during meta-training. This enables MTL2L to update base learners on an unseen dataset at the meta-testing phase. 

\section{Experiments -- Learning to learn image classification} \label{SecExp}
We followed \citet{andrychowicz2016learning} and employed neural optimisers to update MLPs. The neural optimisers were meta-trained to update the MLPs for 100 steps, but during meta-testing, they updated MLPs for the much longer 1000 steps. Base on this setting, we present two experimental scenarios as below.
\begin{itemize}[
noitemsep,topsep=0pt,parsep=0pt,partopsep=0pt,
labelwidth=2cm,align=left,itemindent=1.5cm]
    \item[\textbf{Scenario 1}]: \underline{To test under the common meta-training assumption}\\
    First meta-train neural optimisers to update MLP learners to classify images on one single dataset; and then meta-test by employing the learnt neural optimisers to update MLP learners to classify \textit{the identical dataset} used in meta-training. 
    \item[\textbf{Scenario 2}]: \underline{To test for input-domain heterogeneity}\\
    Unlike \textbf{Scenario 1}, meta-test to update learners to classify \textit{an unseen dataset}.
  \end{itemize}

\newpage
\begin{figure}[ht]
\centering
\begin{subfigure}{0.33\textwidth}
  \centering
  \includegraphics[width=0.975\linewidth]{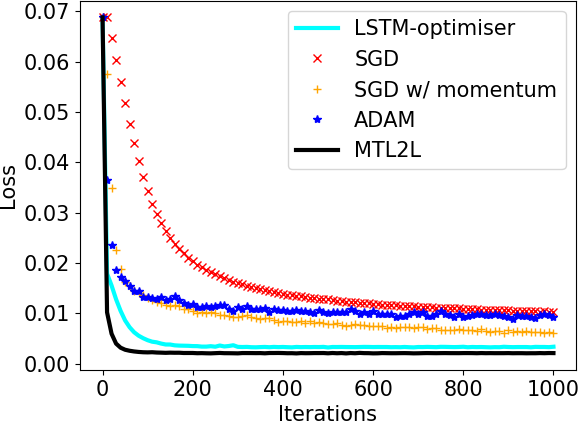}
  \caption{With MNIST}
  \label{fig4a}
\end{subfigure}%
\begin{subfigure}{0.33\textwidth}
  \centering
  \includegraphics[width=0.975\linewidth]{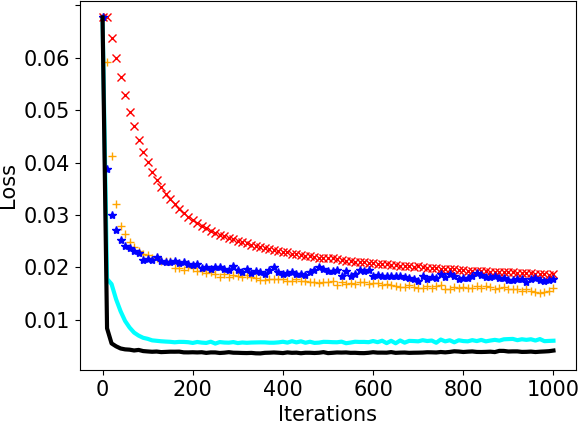}
  \caption{With Fashion-MNSIT}
  \label{fig4b}
\end{subfigure}
\begin{subfigure}{0.33\textwidth}
  \centering
  \includegraphics[width=0.975\linewidth]{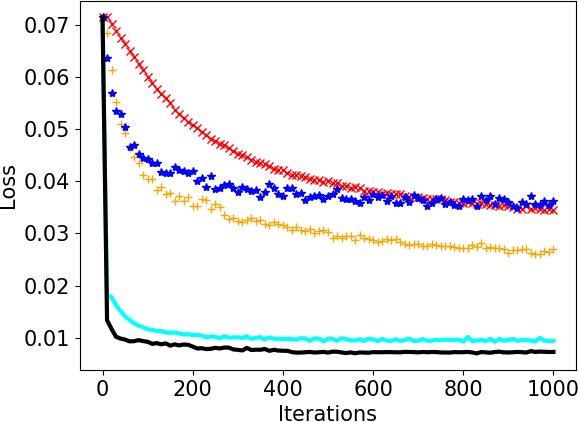}
  \caption{With KMNIST}
  \label{fig4c}
\end{subfigure}
\vspace*{-2.5mm}
\caption{Meta-testing on the same single domain as meta-training}
\label{sec4exp1}
\end{figure}

We tested MTL2Ls against LSTM-optimisers, and against the hand-designed optimisation algorithms of vanilla SGD, SGD with momentum 0.9, and ADAM; all with learning rates 0.01. We presented our results as the learning curves of the learners. The results were colour-coded. We use black solid lines for MTL2Ls, cyan solid lines for LSTM-optimisers, red `$\times$' for vanilla SGD, yellow `$+$' for SGD with momentum, and blue `$\star$' for ADAM.

We upated the MLPs on MNIST~\citep{lecun1998gradient}, Fashion-MNIST~\citep{xiao2017fashion}, KMNIST~\citep{clanuwat2018deep}, and Cifar10~\citep{Krizhevsky2009}. Due to space constraint, we refer the readers to \textbf{Appendix }\hyperref[AppB]{\textbf{B}} for descriptions on the base learner, descriptions on the datasets, hyper-parameters for the neural optimisers, and choices of hypernetworks $\mathscr{N}$s (see \textbf{Section }\hyperref[Sec3]{\textbf{3}}) for MTL2Ls.

The results for \textbf{Scenario 1} are shown in \textbf{Figure }\hyperref[sec4exp1]{\textbf{4}}. For all datasets, the neural optimisers updated learners more rapidly than their handcrafted counterparts. MTL2Ls also converged the learners more efficiently and yielded lower losses than LSTM-optimisers.

\begin{wrapfigure}{r}{.4\textwidth}
\label{sec4fig5}
    \begin{minipage}{\linewidth}
    \centering
    \vspace*{-2mm}
    \includegraphics[width=0.9\linewidth]{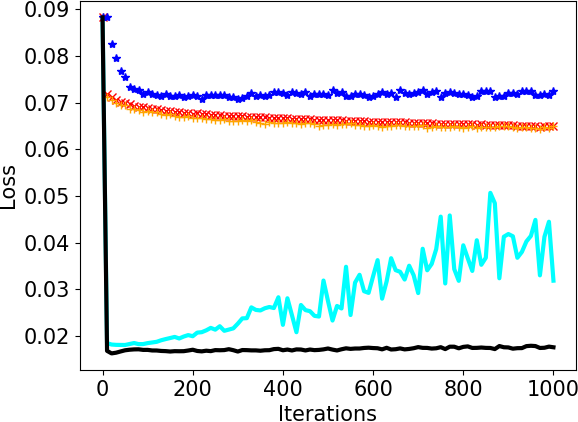}
\end{minipage}
\vspace*{-2mm}
\caption{Meta-tested on an unseen\\ 
\hspace*{16.825mm}Modified Cifar10 dataset}
\vspace{-4mm}
\end{wrapfigure}

The results for \textbf{Scenario 2} are shown in \textbf{Figure }\hyperref[sec4fig5]{\textbf{5}}. Since LSTM-optimisers only supported single-task learning, they were meta-trained to update MLPs for KMNIST. In contrast and as MTL2Ls naturally supports multi-task learning, we meta-trained them to update MLPs on both Fashion-MNIST and KMNIST. During meta-testing, all optimisers updated MLPs on a modified re-sized greyscale Cifar10 dataset (see the synthesis procedure in \textbf{Appendix }\hyperref[AppB2]{\textbf{B.2}}). As shown in the figure, LSTM-optimisers exhibited instability with deteriorating performances whereas MTL2Ls displayed stability. 

It was also interesting to observe the \textit{timing} with which the LSTM-optimisers exhibited instability. As shown in \textbf{Figure }\hyperref[sec4fig5]{\textbf{5}}, the LSTM-optimiser instability was less severe in the first 400 iterations -- the loss curve upraised but with no volatile oscillations. The volatility was only amplified after the first 400 iterations. This showed that there existed some underlying similarities between the task structures for updating learners on KMNIST and on the modified Cifar10. Hence, we conjectured that LSTM-optimisers were actually able to capture the ``overall trends'' in updating learners for an unseen dataset, but with imprecision. Furthermore, such imprecision introduced errors which reiterated in the LSTM and resulted in the eventual deteriorating performances. 

\section{Related Work}\label{sec:5}

In this present work, we extended on the seminal paper of \citet{andrychowicz2016learning} which explored meta-learning via learning to optimise. As stated in Equation \eqref{sec2eq2}, \citet{andrychowicz2016learning} proposed to learn alternative rules to replace the traditional gradient descent methods. Their work has also been extended by the \textit{LSTM-metalearner}~\citep{ravi2016optimization} and \textit{Meta-SGD}~\citet{li2017meta} to learn auxiliary components compatible with the gradient descent regime\footnote{An extended Related Work section can be found in \textbf{Appendix }\hyperref[AppA]{\textbf{A}}.}.

Learning to optimise is one approach to meta-learning. Alternatively, there are metric-based approaches \citep{vinyals2016matching} which learn a kernel function to estimate the similarity between two sampled data. There are also model-based approaches \citep{munkhdalai2017meta} that combine weights updated at different timescales to synchronise different levels of abstracted features.

Besides the task of L2L, there are different ways to formulate the concept of learning quickly. For instance, \textit{few-shot learning}~\citep{vinyals2016matching} trains a network to map unlabelled data to their labels based on a small labelled support set.

\textit{Model-agnostic meta-learning} (MAML)~\citep{finn2017model} is a meta-learning technique that addresses both input-domain heterogeneity and learner architecture heterogeneity (see \textbf{Figure }\hyperref[Fig:Intro-F1]{\textbf{1}}). Instead of training a neural optimiser from the scratch, MAML adopts a double-loop optimisation setup with hand-designed optimisation algorithms. The inner loop finds new candidate weights for the learner, and the outer loop aggregates all candidates to find an optimal update. Hence MAML optimises the parametric initialisation for the least amount of updates to converge a learner.

The \textit{hierarchically structured meta-learning} (HSML)~\citep{pmlr-v97-yao19b} extended on the idea of MAML. HSML jointly trains a task clustering network to learn features specific to datasets as parametric initialisation for the base learner.

\section{Conclusion}
In this paper, we showed that neural optimisers are able to perform beyond the common meta-learning assumption: \textit{to meta-test on the single identical input-domain used during meta-training}. We introduced our novel context aware neural optimiser \textbf{Multi-task Learning to Learn (MTL2L)} with synaptic connections described with SVD and hypernetworks. MTL2Ls encoded input features as eigenvalues for their synapses to self-modify its optimisation rules. In our experiments, we meta-trained our MTL2L on Fashion-MNIST and KMNIST, then meta-tested it on an unseen modified Cifar10 dataset. We demonstrated that MTL2Ls successfully converged learners on an unseen input-domain, thereby making neural optimisers more generally applicable.

\acks{This research was supported by the Australian Government Research Training Program (AGRTP) Scholarship. We also thank our reviewers for the constructive feedback.}
\newpage
\bibliography{main}


\appendix
\hspace{-6mm}
\section{Related Work (part 2)}\label{AppA}

Our focus in \textbf{Section }\hyperref[sec:5]{\textbf{5}} was to highlight different approaches in meta-learning. Here and as requested by our reviewers, we present an additional literature review on studies that extended the original work of \citet{andrychowicz2016learning}.

\subparagraph{Remedies to Low Adaptability in Learnt Optimisers} As mentioned in the \hyperref[sec:1]{\textbf{Introduction}}, it is not well studied whether neural optimisers are capable of updating base learners of a different architecture during the meta-testing phase.

As demonstrated in their own work, \citet{andrychowicz2016learning} showed (in Section 3.2 page 5 of that paper) that a learnt LSTM-optimiser was not generally applicable to changes made in activation functions. They changed the activation function of a MLP base learner from \texttt{tanh} to \texttt{ReLU} and showed that the modified MLP was updated unsuccessfully. This negative result was again brought up by the original panel of authors in \cite{wichrowska2017learned} and was also independently studied by \cite{lv2017learning}. The two mentioned papers addressed different solutions to overcome this problem.

In \cite{lv2017learning}, the authors solved this issue by taking inspiration from the handcrafted adaptive optimisation algorithms. They explicitly modified the input of their RNN-optimiser similarly to ADAM~\citep{kingma2014adam} to allow current derivatives to be reduced slightly through dependencies on past derivatives. Whereas in \cite{wichrowska2017learned}, the authors presented a solution similar to processes of relation classification in text \citep{luo2017recurrent}. They presented a hierarchical multi-layered RNN-optimiser capable of abstracting information of base learner gradients at different levels.

Unlike RNN-optimisers, low adaptability is much harder to be remedied in non-RNN-based learnt optimisers. For instance, Meta-SGD~\citep{li2017meta} trains a unique learning rate per parameter for the base learner. Hence, the amount of parameters of the base learner becomes a hyperparameter of the neural optimiser. It is thus inapplicable to meta-test Meta-SGD on ResNet20s if it was meta-trained to update MLPs.

\subparagraph{Unable to Handle Large Amount of Descent Steps} As mentioned in \textbf{Section }\hyperref[SecExp]{\textbf{4}}, \citet{andrychowicz2016learning} only employed LSTM-optimisers to update MLP base learners for 100 steps. Thus whether LSTM-optimisers are capable to handle large amount of descent steps has been a research topic with much attention in the community. Again, both \cite{lv2017learning} and \cite{wichrowska2017learned} have shown that the vanilla LSTM-optimiser was incapable of updating the base learner for a prolonged duration, and this could be solved by \cite{lv2017learning} through the adaptive gradient modification, and by \cite{wichrowska2017learned} through the hierarchical abstraction architecture. 

A similar study was also conducted for Meta-SGD. \cite{metz2019understanding} analysed the maximum eigenvalue of the objective loss with respect to the meta-learner parameters. They took a variational bound approach to prevent gradient explosion in the training of the meta-learner and remedied biased introduced by truncated backpropagation through time. However, \cite{Kuo_2020_CVPR_Workshops} also showed that, without modification, naive Meta-SGD was able to update base learners up to 5000 steps.

\subparagraph{Other Types of Learnt Optimisers} In this paper, we have mainly focused on learnt supervised learning optimisation algorithms. However, a large amount of research has also been devoted to learn reinforcement learning optimisation algorithms.

For instance, the original panel of authors of \citet{andrychowicz2016learning} have also extended the LSTM-optimiser under the reinforcement learning (RL) setting in \citet{chen2017learning}. That is, they trained an RNN to learn RL. Similarly and as shown in \citet{wang2016learning}, the learning of a RL algorithm could also be learnt by another independent RL algorithm. Their key concept was to use a standard RL to train an RNN which implements its own free-standing RL procedure. The latter idea has also been implemented in an adjacent field of meta-learning. In neural architecture search, \citet{zoph2016neural} have shown that an RNN could be trained with RL to maximised the expected accuracy of the generated architectures on a validation set.

\newpage
\label{AppB}
\section{Experimental Details for Section 4}

This appendix documents the experimental details for \textbf{Section }\hyperref[SecExp]{\textbf{4}}. Information will be provided, in the order of
\begin{itemize}[
noitemsep,topsep=0pt,parsep=0pt,partopsep=0pt,
labelwidth=5mm,align=left,itemindent=0.5cm]
    \item[\textbf{(1)}]: \underline{The MLP Learner,}
    \item[\textbf{(2)}]: \underline{Dataset description,}
    \item[\textbf{(3)}]: \underline{Gradient pre-processing,}
    \item[\textbf{(4)}]: \underline{The meta-training hyper-parametric setups,}
    \item[\textbf{(5)}]: \underline{Choices of orthonormal SVD matrices, hypernetwork $\mathscr{N}$s, and}\\
    \hspace*{8.5mm}\underline{the MTL2L multi-task setup.}
  \end{itemize}
 
\subsection{The MLP Learner}\label{AppB1}
Our MLP learners were simple. They consisted one linear layer followed by a ReLU activation, and one softmax layer. The linear layer has input dimension $784$, the softmax layer has output dimension $10$, and the hidden dimension connecting the linear layer and softmax layer was set to $32$. The specific reasons for numbers $784$ and $10$ will be made clear in the \hyperref[AppB1]{\textbf{next section}} for data description; and the MLP learners had $25.4$K parameters.

\subsection{Dataset description}\label{AppB2}
We upated the MLP learners on MNIST~\citep{lecun1998gradient}, Fashion-MNIST~\citep{xiao2017fashion}, KMNIST~\citep{clanuwat2018deep}, and Cifar10~\citep{Krizhevsky2009}. We will first discuss the three MNIST datasets, and then separately address Cifar10.

The MNIST dataset contains 60,000 training and 10,000 test images in 10 classes of 28$\times$28 greyscale handwritten digits. This is a popular dataset among the machine learning community and commonly used as a benchmark to validate their algorithms. The subsequent datasets of Fashion-MNIST and KMNIST were crafted in a similar style to that of the original MNIST dataset. Both Fashion-MNIST and KMNIST has 60,000 training and 10,000 test greyscale images with 10 classes of items on 28$\times$28 pixels. While Fashion-MNIST consists iamges of fashion items, KMNIST consists kuzushiji (\hspace*{-3mm}
\begin{CJK}{UTF8}{min}
崩し字
\end{CJK}\hspace{-1mm}) versions of hiragana (\hspace*{-3mm}
\begin{CJK}{UTF8}{min}
平仮名
\end{CJK}\hspace{-1mm}) characters from classical Japanese literature.

We choose MNIST, Fashion-MNIST, and KMNIST for the similarities in their dataset constructions. All images were presented as a vector of $784(= 28 \times 28)$ pixels to the MLP learners. Hence the input dimension of the MLPs was $784$, and the output dimension was $10$ due to the amount of labels. 

The CIFAR-10 dataset is slightly more challanging. It consists 32$\times$32 RGB-colour images in 10 classes, and there are 50,000 training images and 10,000 test images. In order to fit these images in the MLP learner of \textbf{Appendix }\hyperref[AppB1]{\textbf{B.1}}, we created a synthesised Cifar10 dataset consisting of greyscale 28$\times$28 pixel images. The images were made greyscale by combining the coloured filters in the ratio of $0.30 \text{R} + 0.59 \text{B} + 0.11 \text{G}$. Furthermore, the images were made 28$\times$28 by only selecting the first 28 rows and the first 28 columns.

\subsection{Gradient pre-processing}\label{AppB3}

This subsection reiterates the following content from \citet{andrychowicz2016learning} for the completeness of this paper. The original content can be found on page 11 of their paper under ``A Gradient preprocessing''.

\citet{andrychowicz2016learning} proposed to pre-process the neural optimiser’s input gradient $\nabla_\theta \mathcal{L}_{t+1}(\theta_t)$ (which we simplify as $\nabla$ in this subsection) with

$\nabla^{(j)} \rightarrow $\hspace{3mm} $
\left\{
                \begin{array}{ll}
                  \left(\frac{\text{log}(| \nabla |)}{p}, \text{sgn}(\nabla)\right) \hspace{5mm} \text{if } |\nabla| \ge e^{-p},\\
                  (-1, e^{p}\nabla) \hspace{16.75mm} \text{otherwise .} 
                \end{array}
              \right.$\\
              
\hspace{-6mm}Each element of the native gradient $\nabla^{(j)}$ for $j = 1 \ldots \lvert\theta_t\lvert$ is pre-processed as a pair of values. The hyper-parameter $p$ controls how small gradients are disregarded, and defaults to $10$ in all of their and in all of our experiments. This scheme was devised to ensure that the magnitudes of every dimension is on the same order. \citet{andrychowicz2016learning} mentioned that this is necessary because neural networks, and including neural optimisers ``\textit{naturally disregard small variations in input signals and concentrate on bigger input values}''.

\subsection{The meta-training hyper-parametric setups}\label{AppB4}
The RNNs of the LSTM-optimisers and MTL2Ls were formulated as documented in \textbf{Table }\hyperref[sec3tab1]{\textbf{1}}; and both were trained with the ADAM optimiser with learning rate 0.001. Following \citet{andrychowicz2016learning}, we set the RNNs as 2-layer deep with hidden dimension 20. The inputs to the RNNs were the pre-processed gradients of \textbf{Appendix }\hyperref[AppB3]{\textbf{B.3}}; they were treated as a data of batch size $\lvert\theta_t\lvert$ with feature size (input dimension) 2. Due to this reason, the input dimension of the RNNs of both the LSTM-optimisers and MTL2Ls were of size 2. Below, we applied the additional hyper-parametric setups to \textbf{Algorithm }\hyperref[sec2alg1]{\textbf{1}} for meta-training. 

For all LSTM-optimisers, we set \colorbox{Xi-col}{$\Xi$}$ = 5$, \colorbox{Q-col}{$\mathcal{Q}$}$ = 20$, and \colorbox{S-col}{$\mathcal{S}$}$ = 100$. The combination of these hyper-parameters meant that, LSTM-optimisers were trained to update \colorbox{Q-col}{$20$} MLP learners for \colorbox{S-col}{$100$} steps; and that the LSTMs were unrolled every 5 steps for updates. Thus they memorised \colorbox{Xi-col}{$5$} continual steps of the optimisation trajectory for the MLP learners. 

MTL2Ls of experimental \textbf{Scenario 1} were setup identically as the LSTM-optimisers. MT2Ls of experimental \textbf{Scenario 2} were setup differently to enable multi-task learning. For \textbf{Scenario 2}, we set \colorbox{Xi-col}{$\Xi$}$ = 5$, \colorbox{Q-col}{$\mathcal{Q}$}$ = 30$, and \colorbox{S-col}{$\mathcal{S}$}$ = 100$. It will be made clear in \textbf{Appendix }\hyperref[AppB5]{\textbf{B.5}} for the reason of the prolonged learner-trial \colorbox{Q-col}{$\mathcal{Q}$}.

\subsection{Choices of orthonormal SVD matrices, hypernetwork $\mathscr{N}$s, \\\hspace{9mm}and the MTL2L multi-task setup}\label{AppB5}
The main novelty in our MTL2L neural optimiser are the self-modifying synapses in Equations \eqref{MTL2L_W} and \eqref{MTL2L_U}. In this subsection, we addresses how to configure the MTL2L SVD synapses and setup the MTL2L multi-task learning. We will discuss in the order of, the orthonormal SVD matrix selections, the hypernetwork $\mathscr{N}$s for experimental \textbf{Scenarios 1} and \textbf{2}, and then the MTL2L multi-task learning setups for \textbf{Scenario 2}.

\subsubsection{Orthonormal matrix selection for SVD synapses}
Equations \eqref{MTL2L_W} and \eqref{MTL2L_U} describe the MTL2L synapses with SVD. Abiding to the SVD formulation, the purple components are orthonormal matrices, whereas the orange components are diagonal matrices with eigenvalues (or singular values, to be more precise). We emphasise again that the purple matrices are fixed prior to training, and only the orange components are trained with hypernetworks. More on the orange components will be addressed in the following subsubsections.

Any means of orthonormal matrix generation is viable. The authors of this manuscript coded in Python~\citep{CS-R9526}, and we choose to use the SciPy library~\citep{2020SciPy-NMeth}. Refer to \url{https://docs.scipy.org/doc/scipy/reference/generated/scipy.stats.ortho_group.html} for guidelines.

\subsubsection{For Scenario 1}
The hypernetwork $\mathscr{N}$s for \textbf{Scenario 1} are MLPs. These hypernetwork-MLPs consisted two linear layers -- the first linear layer was followed by a ReLU activation, then followed by the second linear layer. The hypernetwork-MLPs received the images of as vectors of $784$ pixels. Hence, the input dimension of the first linear layer was set to $784$. In addition, the output dimension of the second linear layer was $20$. This is because the hypernetwork-MLPs were employed to infer for the eigenvalues of the SVD synapses of MTL2Ls. Hence, the output dimension of the second linear layer was required to be identical to the hidden dimension of the MTL2Ls. Last, the hidden dimension which connects the first and the second linear layer of the hypernetwork-MLPs were set to 32.

\subsubsection{For Scenario 2}
As mentioned in \textbf{Section }\hyperref[SecExp]{\textbf{4}}, we set MTL2Ls to perform multi-task learning on Fashion-MNIST and on KMNIST for the meta-training phase of experimental \textbf{Scenario 2}. As discussed in \textbf{Appendix }\hyperref[AppB4]{\textbf{B.4}}, the learner-trial \colorbox{Q-col}{$\mathcal{Q}$} was set as the slightly longer 30 -- 15 trials were set to update the MLP learners for the Fashion-MNIST dataset, and the remaining 15 were reserved to update the MLP learners on KMNIST.

During meta-training, we alternated between updating the MLP learners for Fashion-MNIST and for KMNIST as illustrated in \textbf{Figure }\hyperref[sec2fig3]{\textbf{3(b)}}. That is, when \colorbox{Q-col}{$\mathcal{Q}$}$ = 1, 3, \ldots, 29$, MTL2Ls updated MLP learners to classify on Fashion-MNIST; while for even-number \colorbox{Q-col}{$\mathcal{Q}$}s, MTL2Ls updated MLP learners to classify on KMNIST.

The hypernetwork $\mathscr{N}$s for \textbf{Scenario 2} were similar to those in \textbf{Scenario 1} but with a slight modification. We formulated them as 
\begin{equation}
    \mathscr{N}_t = 0.9 \mathscr{N}_{t-1} + 0.1 \text{MLP}(\mathbf{x}_t), \text{with} \label{eq_fin}
\end{equation}
$\mathscr{N}_0 = \vec{\textbf{0}}\in\mathbb{R}^{20}$. The hypernetwork-MLPs employed in Equation \eqref{eq_fin} were identical to those employed in \textbf{Scenario 1}. However in \textbf{Scenario 2}, we treated the propagation of the eigenvalues of the SVD synapses of MTL2Ls like momentum~\citep{nesterov1983method}. This was done to increase training stability for MTL2L.

As a final and less significant note, we injected small-sized perturbations to the parameters of the MLP learners during the meta-training phase. This is done to incorporate more noise during the training of MTL2Ls. 

\end{document}